\documentclass[letterpaper, 10 pt, conference]{ieeeconf}
\IEEEoverridecommandlockouts

\usepackage[percent]{overpic}

\usepackage{tabu}

\usepackage{multirow}
\def\mr{\multirow}
\def\mc{\multicolumn}

\usepackage[overload]{empheq}             

\usepackage[table,xcdraw]{xcolor}
\definecolor{dg}{rgb}{0.1, 0.6, 0.2}       
\definecolor{b}{rgb}{0.0, 0.0, 1}          
\usepackage{colortbl}                      

\usepackage{epsfig}
\usepackage{float}
\usepackage{color}
\usepackage{url}

\usepackage{algpseudocode}
\usepackage[ruled,linesnumbered,vlined]{algorithm2e}

\usepackage{amssymb, amsfonts}
\usepackage{amsmath}
\usepackage{mathrsfs}

\usepackage{amsthm}
\usepackage{mathtools}

\let\labelindent\relax
\usepackage{enumitem}       
\setenumerate[enumerate]{align=left}

\usepackage{graphicx}
\usepackage{subcaption}
\usepackage{caption}

\usepackage{cite}

\makeatletter
\let\NAT@parse\undefined
\makeatother
\usepackage{hyperref} 

\usepackage{balance}

\usepackage{bm}

\usepackage{rotating}

\newcommand{\norm}[1]{\left\lVert#1\right\rVert}
\newcommand{\abs}[1]{\left\lvert#1\right\rvert}

\newcommand{\Mint}{%
  \mathrlap{\mathop{\phantom{\int}}\limits_{\!\mathcal{Q}}}%
  \!\int
}

\makeatletter
\newlength\tmp@\newlength\t@mp
\newcommand{\comp}[3]
  {\mathop{ \settowidth\tmp@{$\displaystyle\mathop{#1}^{#3}_{#2}$}
  \hbox to \tmp@{\hss \settowidth\t@mp{$\displaystyle #1$}\setlength\t@mp{.45\t@mp}
  $\displaystyle\mathop{#1}^{\hspace\t@mp #3}_{\hspace{-\t@mp}#2}$
  \hss} }}
\makeatother

\newcommand{\Int}[2]
{\int_{#1}^{#2}}

\DeclareMathOperator*{\argmin}{argmin}

\hypersetup{
    colorlinks=true,
    linkcolor=blue,
    filecolor=blue,      
    urlcolor=blue,
    citecolor=blue,
}

\def\a{\alpha}
\def\b{\beta}
\def\d{\delta}

\def\g{\gamma}

\def\o{\omega}
\def\s{\sigma}

\def\D{\Delta}

\def\R{\mathbb{R}}
\def\N{\mathbb{N}}

\def\l{\left}
\def\r{\right}

\def\F{\vbf{F}}
\def\M{\vbf{M}}
\def\Fcoef{\mathcal{F}}

\def\f{\mathbf{f}}

\def\quat{\mathbf{q}}

\def\pos{\mathbf{p}}
\def\vel{\mathbf{v}}
\def\accel{\mathbf{a}}

\newcommand{\fr}[1]{\texttt{#1}}

\newcommand{\vbf}[1]{\bm{\mathbf{#1}}}

\def\resi{{r}}
\def\L{\mathcal{L}}

\def\bias{\mathbf{b}}

\def\rot{\mathbf{R}}
\def\tf{\mathbf{T}}

\def\SO{\mathrm{SO(3)}}

\def\SE{\mathrm{SE(3)}}

\def\qtoR{\mathcal{R}}
\def\Exp{\mathrm{Exp}}
\def\Log{\mathrm{Log}}

\def\zero{\mathbf{0}}

\def\n{\vbf{n}}
\def\dis{\vbf{d}}

\def\Dt{ {\D t} }
\def\dt{ {\d t} }
\def\wrt{\text{w.r.t. }}

\def\pia{{\vbf{\a}}}
\def\pib{{\vbf{\b}}}
\def\pig{{\vbf{\g}}}
\def\pith{{\vbf{\theta}}}

\def\X{\mathcal{X}}

\def\U{\mathcal{U}}
\def\I{\mathcal{I}}

\def\angvel{\bm{\omega}}
\def\accel{\mathbf{a}}
\def\grav{\mathbf{g}}

\def\fA{\fr{A}}
\def\fB{\fr{B}}

\def\fW{\fr{W}}

\begin{document}

\title{\bf LIRO: Tightly Coupled Lidar-Inertia-Ranging Odometry}

\author{
      Thien-Minh Nguyen,
      Muqing Cao,
      Shenghai Yuan,
      Yang Lyu,
      Thien Hoang Nguyen,
      and Lihua Xie 
\thanks{
The work is supported by National Research Foundation (NRF) Singapore, ST Engineering-NTU Corporate Lab under its NRF Corporate Lab@ University Scheme.}
\thanks{The authors are with School of Electrical and Electronic Engineering, Nanyang Technological University, Singapore 639798, 50 Nanyang Avenue (e-mail: thienminh.npn@ieee.org)
}
}

\markboth{
    Confidential - Original \& Unpublished Manuscript
    }
{Nguyen \MakeLowercase{\textit{et al.}}: Visual-Inertial-Ranging-Aided Localization via Optimization-Based Sensor Fusion Approach}

\maketitle

\thispagestyle{plain}
\pagestyle{plain}

\begin{abstract}
In recent years, thanks to the continuously reduced cost and weight of 3D Lidar, the applications of this type of sensor in robotics community have become increasingly popular. Despite many progresses, estimation drift and tracking loss are still prevalent concerns associated with these systems. However, in theory these issues can be resolved with the use of some observations to fixed landmarks in the environments. This motivates us to investigate a tightly coupled sensor fusion scheme of Ultra-Wideband (UWB) range measurements with Lidar and inertia measurements. First, data from IMU, Lidar and UWB are associated with the robot's states on a sliding windows based on their timestamps. Then, we construct a cost function comprising of factors from UWB, Lidar and IMU preintegration measurements. Finally an optimization process is carried out to estimate the robot's position and orientation. It is demonstrated through some real world experiments that the method can effectively resolve the drift issue, while only requiring two or three anchors deployed in the environment.
\end{abstract}



\section{Introduction} \label{sec: intro}
Localization is a crucial task that requires a lot of attention and effort in mobile robotics. Depending on the scenario, one has has to factor in the trade-offs among cost, accuracy, robustness, computational load, and ease of deployment to settle on the most appropriate method for the robot at hand.
For most applications, selecting the set of sensors is usually the first concern that has to be addressed, especially in GPS-denied environments.
In some cases, one can employ some artificially installed sensors such as motion-capture \cite{kushleyev2013towards, watterson2020control}, visual markers \cite{martinez2009trinocular, klose2010markerless}, or Ultra-wideband (UWB) beacons \cite{mueller2015fusing, guo2016ultra, nguyen2016ultra, tiemann2017scalable, paredes20183d}. On the other hand, to operate in complex and uncontrolled GPS-denied environments, Onboard Self-Localization (OSL) methods such as visual-inertial odometry (VIO) \cite{forster2014svo, shen2015tightly, qin2017vins, bloesch2017iterated, engel2014lsd, mur2017orb, loianno2017estimation, weinstein2018visual} or Lidar Odometry and Mapping (LOAM) \cite{zhang2018laser, ye2019tightly, zuo2019lic, chen2020sloam} techniques are often the most viable option for robot localization. Indeed, it has become increasingly clear that OSL systems are going to be the backbone of autonomous navigation for the years to come.

Among OSL methods, 3D Lidar based approaches are gaining more and more popularity thanks to the continuously reduced cost and weight of this type of sensor in recent years. Originally weighing over 10 kg and costing tens of thousands of USDs,
Lidar could only be used in large autonomous cars. In contrast, currently some commercial products only weigh for a few hundred grams and cost only a few hundred USDs.
Even for Micro/Unmanned Aerial Vechicles (MAVs / UAVs), where the payload capacity is limited, successful uses of Lidar have been demonstrated in recent years \cite{zhang2018laser, nguyen2020autonomous, dharmadhikari2020motion, reinhart2020learning}. Compared to common camera-based systems, a Lidar-based solution offers several advantages. For instance, Lidar can directly observe the metric scale features, hence, it can directly provide self-localization information (odometry) for the robot, while simple monocular camera-based solutions only provide odometry data of ambiguous scale. Moreover, even when compared with RGDB or stereo cameras, which are on the same footing with Lidar as they can detect metric-scale features, Lidar has a much higher sensing range, besides being almost invariant to illumination conditions.

Despite the above advantages, LOAM, being an OSL method, is still prone to estimation drift and loss of tracking due to lack of features in an environment. In addition, under the OSL approach, robots can only estimate their positions relative to their initial location. These issues can be quite inconvenient in applications such as inspection of 3D structures, where predefined trajectories are often desired to be executed in reference to a chosen so-called world frame. One solution can be to fuse GPS poses with LOAM \cite{shan2020liosam}, however it can only be effective in wide open areas, since GPS will strongly degrade in clustered and dense urban environments. Another approach could be to train a neural network on recognizing segmented objects from the pointcloud for place recognition and loop closure \cite{tinchev2019learning}. Obviously this approach also requires significant effort in collecting data, labelling and training, not to mention that future changes in the environment can cause the system to lose effectiveness.

In this paper, we posit that by using ranging measurements from the robot to two or three UWB anchors, which can be easily deployed in the environment, long-term drift in both position and orientation can be effectively eliminated. Moreover, the pose (i.e. both position and orientation) estimate can also be referenced in the desired coordinate system defined via the anchor deployment. In this case, we say that our Lidar-based odometry estimate is \textit{drift-free} and \textit{global}. The main contributions of this work can be listed as follows:

\begin{itemize}
    \item We integrate a flexible ranging scheme with a Lidar-inertial odometry system to achieve drift-free and global pose estimates.
    \item We develop an efficient tightly coupled sensor fusion framework to combine the so-called body-offset ranges, IMU preintegration, and Lidar-based features for real time estimation of the robot states.
    \item We extensively validate the estimation scheme via real world experiments in practical scenarios.
\end{itemize}

The remaining of this paper is organized as follows: in Section \ref{sec: related works}, we review some related works to highlight the novelty of our approach; Section \ref{sec: preliminary} introduces some preliminaries. Section \ref{sec: framework} presents an overview of the methodology. Section \ref{sec: cost factor} provides a more detailed description of the cost factors and Section \ref{sec: experiment} presents our experiment results. Section \ref{sec: conclusion} concludes our work.

\section{Related Works} \label{sec: related works}

Indeed, in recent years, many researchers have employed ranging to UWB anchors to handle the estimation drift of OSL methods. For example, in \cite{wang2017ultra}, a loosely coupled sensor fusion scheme was proposed to leverage UWB ranges in correcting a pointcloud-based SLAM system. As the term loosely coupled suggests, this scheme fuses the localization information from two relatively independent systems. As such, if there are only two or three anchors, the UWB-based localization thread may not be realizable. We also note that since there is no observation to connect the consecutive poses in the sliding window, this method employs so-called \textit{smoothness factors}, which are based on the maximum velocity between consecutive ranging times. Thus when the real velocity actually exceeds the assumed maximum velocity, the estimate can exhibit "lagging" behaviour of a low-pass filter. This can be resolved with the use of IMU preintegration, which is one of the key techniques featured in our method.

In \cite{nguyen2019loosely}, a loosely coupled method was developed to correct the scale of a monocular visual odometry (VO) output using the distance measurements from a single UWB anchor. In this method the anchor's position is also estimated with respect to the coordinate frame of the VO system, whose origin coincides with the initial position of the camera. Hence the estimate is still of local type (it will change when the VO system initiates at a new position). In \cite{nguyen2019tightly, nguyen2020tightly}, a tightly coupled scheme was investigated, where UWB ranging and ORB features are combined into a cost function, which is then optimized to provide a metric-scaled position estimate. This tightly coupled scheme requires a number of anchors less than four. Note that the estimate is still of local type. In \cite{cao2020vir}, the authors proposed a tightly coupled visual-inertial-ranging sensor fusion scheme. Only one anchor is used in this case, but multiple robots can also exchange range to jointly estimate their relative position. However, we note that this system still has to employ smoothness factors, along with being of local type. Finally, in \cite{song2019uwb}, a loosely coupled approach was proposed to combine 2D lidar odometry estimate with UWB ranges to jointly localize the robot and the UWB anchor positions. We consider that this method is of global type, though the loosely coupled approach would require large number of UWB anchors, besides only providing 2D localization estimate. We also note that all of the aforementioned works only focused on the the position estimation, while orientation state was ignored. This is expected since the UWBs ranging were only conducted in a body-centered manner.

To the best of our knowledge, our work is the first that investigates the tightly coupled sensor fusion approach of 3D Lidar, IMU and UWB ranges (hence the acronym LIRO for \underline{L}idar-\underline{I}nertia-\underline{R}anging \underline{O}dometry). Moreover, we also employ a novel body-offset ranging model that couples the position, orientation and velocity in the range observations. In addition, by using the communication capability of the UWB, and a very simple anchor deployment scheme, we can straightforwardly estimate the anchor position and start the localization in a user-defined frame, instead of having to estimate the anchor positions in a random frame by using VO data. This makes the LIRO scheme a flexible and accurate solution for localization of robot in inspection tasks.

\section{Preliminaries} \label{sec: preliminary}

\subsection{Notations}

In this paper we use $(\cdot)^\top$ to denote the transpose of a vector or matrix under $(\cdot)$.
For a vector $ \vbf{x} \in \R^m $, $\norm{\vbf{x}}$ stands for its Euclidean norm, and $\norm{\vbf{x}}_{\vbf{G}}^2$ is  short-hand for $\norm{\vbf{x}}_{\vbf{G}}^2 = \vbf{x}^\top \vbf{G} \vbf{x}$. For two vectors $\vel_1$, $\vel_2$, $\vel_1\times\vel_2$ denotes their cross product. In the sequel, we shall denote $\rot \in \SO$ as the rotation matrix, and $\tf \in \SE$ as the transformation matrix. $\Log(\rot)$ returns the rotation vector of $\rot$. For a unit quaternion $\quat$, $\qtoR(\quat)$ is its corresponding rotation matrix, and $\vbf{vec}(\quat)$ returns its vector part.

To make explicit that a vector $\vel$, or points in a pointcloud $\F$, are \wrt a coordinate frame $\{\fr{A}\}$, we attach a left superscript ${}^{\fr{A}}$ to $\vel$ or $\F$, e.g. ${}^\fr{A}\mathbf{v}$, ${}^{\fA}\F$. A rotation matrix and transformation matrix between two reference frames are denoted with the frames attached as the left-hand-side superscript and subscript, e.g. ${}^\fA_\fB\rot$ and ${^\fA_\fB\tf}$ are the rotation and transform matrices from frame $\{\fA\}$ to $\{\fB\}$, respectively. When the coordinate frames are the body frame at different time instances, we may also ignore the superscript and subscripts, e.g. ${}^k\rot_{k+1} \triangleq {}^{\fB_k}_{\fB_{k+1}}\rot$, or ${}^w_m\tf \triangleq {}^{\fB_w}_{\fB_{m}}\tf$.
For a list of vectors $\vel_1, \vel_2 \dots, \vel_n$ (including scalar) we may write $(\vel_1, \vel_2 \dots, \vel_n)$ as a short-hand for $[\vel_1^\top, \vel_2^\top \dots, \vel_n^\top]^\top$.

\subsection{State estimates}

We define the robot states to be estimated at time $t_k$ as:
\begin{align}
    \X_k &= \Big( \quat_k,\ \pos_k,\ \vel_k,\ \bias^{\o}_{k},\ \bias^{a}_{k}\Big), \label{eq: state vector}
\end{align}
where $\quat_k$, $\pos_k$, $\vel_k \in \R^3$ are respectively the orientation quaternion, position and velocity \wrt the world frame $\{\fW\}$ at time $t_k$; $\bias^{a}_{k},\ \bias^{\o}_{k} \in \R^3$ are respectively the IMU accelerometer and gyroscope biases. Note that the world frame $\{\fW\}$ is defined by the user through the deployment of the anchors, as explained in Section \ref{sec: ranging scheme}.
We denote the state estimate at each time step $k$, and the sliding windows as follows:
\begin{alignat}{2}
    &\hat{\X}_k &&= \Big( \hat{\quat}_k,\ \hat{\pos}_k,\ \hat{\vel}_k,\ \hat{\bias}^{\o}_{k},\ \hat{\bias}^{a}_{k}\Big), \label{eq: X hat k}\\
    &\hat{\X} &&= \l(\hat{\X}_k,\ \hat{\X}_{k+1} \dots\ \hat{\X}_{k+M}\r), \label{eq: X hat k to k+M}
\end{alignat}
where $M \in \N$ is the number of steps in the sliding windows. We choose $M = 10$ for all of the experiments in this work.

Note that in this work the extrinsic parameters have been manually calibrated and set as prior. Indeed, we find introducing these extra states to the problem yields inconclusive benefit while also slowing down the convergence rate.

\section{Lidar-Inertial-Ranging Odometry Framework} \label{sec: framework}

\begin{figure}
    \setlength\belowcaptionskip{-0.25cm}
    \centering
    \begin{overpic}[width=\linewidth,
                        ]{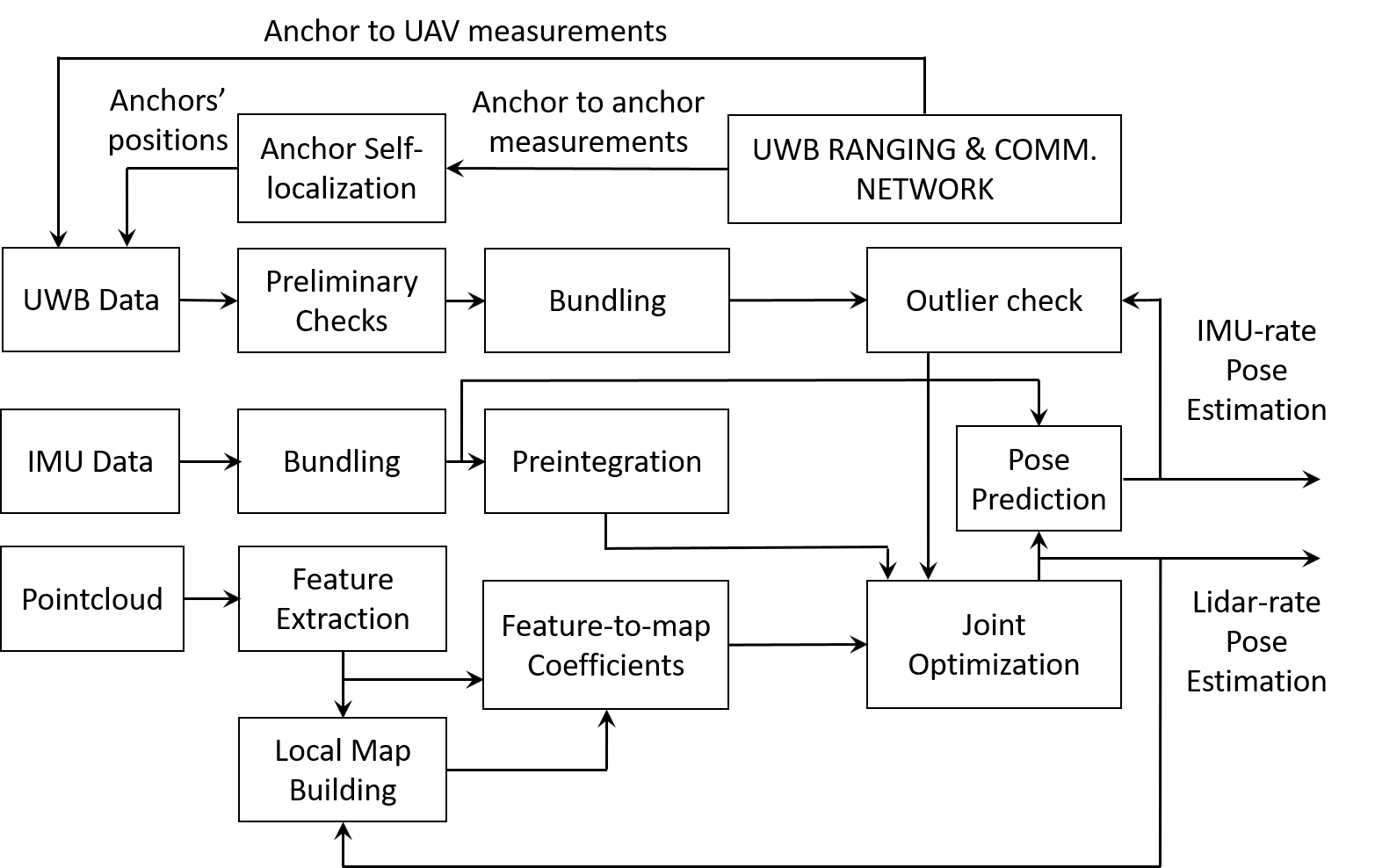}
                        \put(59.5,  30){\footnotesize  $\{\U^i_k\}$}
                        \put(55.0,  25){\footnotesize  $\I_k$}
                        \put(53.5,  18){\footnotesize  $\{\L^i_m\}$}
                        
                        \put(45.00, 8.0){\footnotesize $\M_{w}$}
                        \put(18.00, 12.0){\footnotesize $\F_{m}$}
                        \put(40.00, 2.5){\footnotesize $\{\hat{\tf}_{w}, \hat{\tf}_{w+1}, \dots \hat{\tf}_{k} \}$}
	\end{overpic}
	\caption{The Lidar-Inertia-Ranging Odometry framework. A snapshot of the main quantities at time $t_{k+1}$ when a new pointcloud message just arrives is ovelaid on the connections.}
	\label{fig: block diagram}
\end{figure}
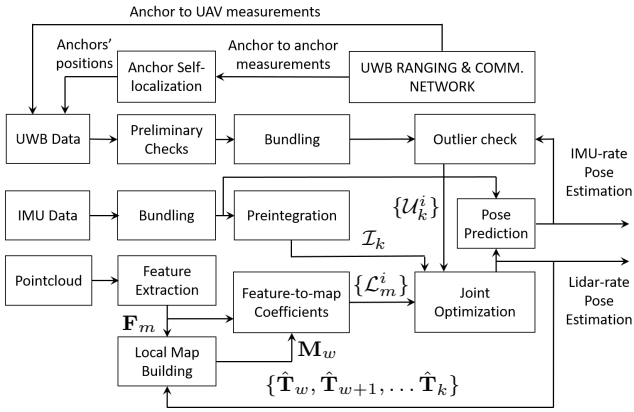

Fig. \ref{fig: block diagram} provides an overview of our framework, where each main process will be described in more details below.

\subsection{UWB ranging and communication network} 
\label{sec: ranging scheme}

\begin{figure}
    \setlength\belowcaptionskip{-0.25cm}
    \centering
    \begin{overpic}[width=\linewidth,
                        ]{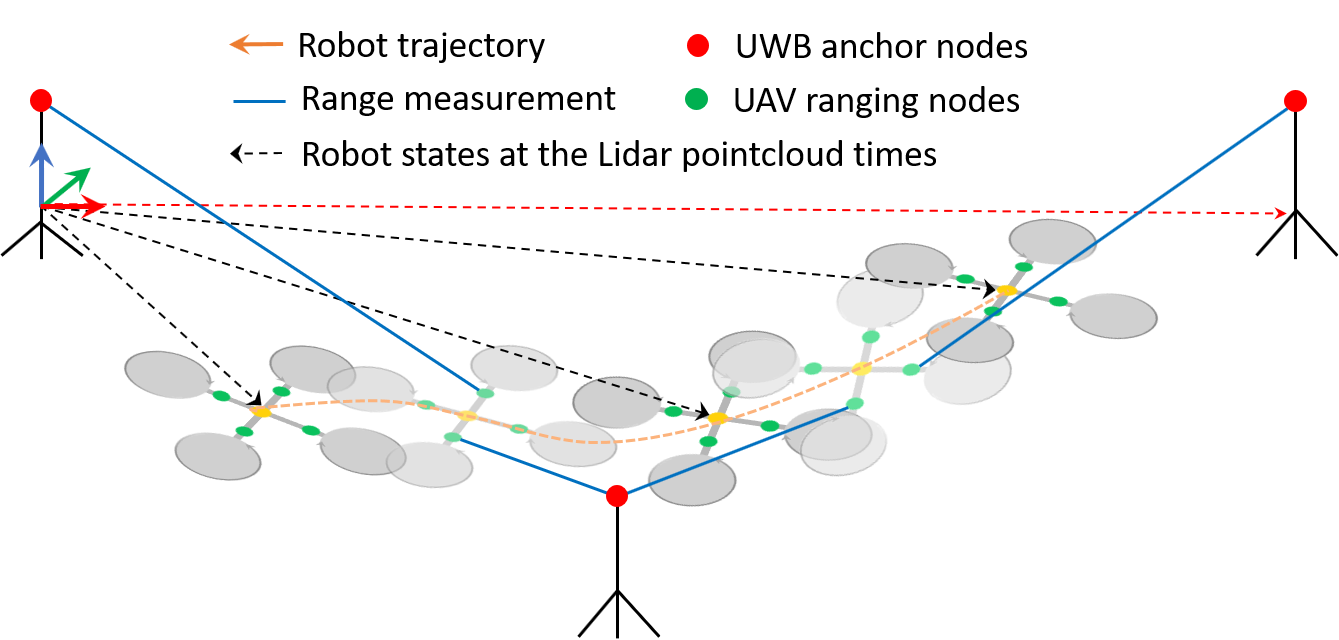}
		\put(01.0,  25.50){\footnotesize $\{\fW\}$}
        \put(15.00, 23.00){\footnotesize $\X_k$}
        \put(40.00, 22.50){\footnotesize $\X_{k+1}$}
        \put(55.00, 29.00){\footnotesize $\X_{k+2}$}
        
        \definecolor{distancolor}{RGB}{0, 112, 192}
        \put(26.00, 26.00){\footnotesize \textcolor{distancolor}{$\breve{d}_{k}^1$}}
        \put(37.50, 09.00){\footnotesize \textcolor{distancolor}{$\breve{d}_{k}^2$}}
        \put(56.50, 10.50){\footnotesize \textcolor{distancolor}{$\breve{d}_{k+1}^1$}}
        \put(80.00, 35.00){\footnotesize \textcolor{distancolor}{$\breve{d}_{k+1}^2$}}
        \definecolor{anchorcolor}{RGB}{112, 48, 160}
        
        \put(00.50, 43.00){\footnotesize \textcolor{anchorcolor}{$(0, 0, z^*)$}}
        \put(87.00, 43.00){\footnotesize \textcolor{anchorcolor}{$(x^*_1, 0, z^*)$}}
        \put(47.00, 06.00){\footnotesize \textcolor{anchorcolor}{$(x^*_2, y^*_2, z^*)$}}
        
	\end{overpic}
	\caption{Illustration of the ranging scheme over three time steps. 
	Note that the time index $k$ of the distance sample refers to the period $(t_k, t_k]$ in which it is obtained, not the exact time instance $t_k$.}
	\label{fig: ranging scheme}
\end{figure}
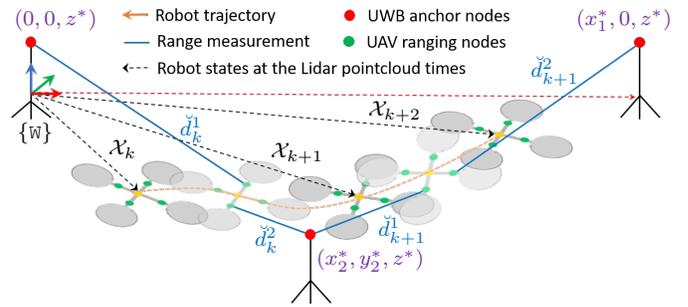

At the top of Fig. \ref{fig: block diagram}, we can see a UWB ranging and communication network, which allows us to measure the distances between the anchor nodes, and between anchor nodes and the ranging nodes on the robot.
Fig. \ref{fig: ranging scheme} illustrates the ranging scheme in more details. First, we have a set of anchors deployed in the area of operation as fixed landmarks. To keep this task simple, we only focus on the scenarios with two or three anchors having the same height from the ground, denoted as $z^*$. It can be seen that two such anchors are already sufficient to define a coordinate system. Specifically, anchor 0 can be assumed to be at the $(0, 0, z^*)$ position, while the direction from anchor 0 to anchor 1 shall define the $+x$ direction. Hence, anchor 1's coordinate can set as $(x^*_1, 0, z^*)$, where $x^*_1$ is obtained by having anchor 1 directly range to anchor 0 multiple times, broadcast these measurements to the robot, and take the average as $x^*_1$. The position of the third anchor can also be calculated by simple geometry. Note that ranging and communicating capabilities are readily supported by the UWB nodes used in this work.

The next feature in the ranging scheme is the UWB ranging nodes on the robot. In this work, multiple UWB nodes are installed on the robot, whose position in the robot's body frame is known. This body-offset ranging scheme is an important feature that allows the ranges measurements to be coupled with the robot orientation, hence allowing the orientation estimate to be of global type.


\subsection{UWB measurement workflow}

After the anchor positions have been determined, the robot-to-anchor ranges can be used for the estimation process.
First, they are put through some preliminary checks based on the signal over noise ratio, line-of-sight indicator, rate of change, etc. to remove any obvious unreliable measurements. The measurements are then stored in a buffer. Now, assuming that the system is at the time step $t_{k+1}$, which corresponds to the arrival of a new pointcloud message and the creation of the state $\hat{\X}_{k+1}$, the UWB buffer is then checked and all measurements that arrived during the period $(t_{k}, t_{k+1}]$ will be "bundled" together as a set. This set will be checked again using the latest IMU-predicted states to remove the suspected outliers. The final result is the set of $N^k_\U$ UWB measurements obtained during $(t_{k}, t_{k+1}]$, denoted as $\{\U^i_k : i = 1, 2,\dots N^k_\U\}$.

\subsection{IMU measurement workflow}

The workflow of the IMU measurement is simpler, where we also extract from the buffer the IMU measurements that arrive during the period $(t_{k}, t_{k+1}]$ when the time step $t_{k+1}$ elapses. These measurements are then preintegrated to provide a single IMU preintegration measurement $\I_k$ that couples two consecutive states $\X_{k}$ and $\X_{k+1}$. Also, IMU measurements are also used to propagate the robot state from the last jointly optimized estimate.
For example, at time $t_{k+1}$ where we have just received a new pointcloud message, $\hat{\tf}_k$ is the last joint-optimization-based estimate of the robot pose.

\subsection{Lidar pointcloud workflow}

The handling of Lidar pointclouds are done similarly to the LIO-mapping framework developed in \cite{ye2019tightly}. Upon receiving a complete sensor scan at time $t_{k+1}$, the features can be extracted to form a feature pointcloud $\F_{k+1} \triangleq {}^{\fB_{k+1}}\F_{k+1}$ (which is actually a composite of two pointclouds, one consists of plane features, denoted as $\F^p_{k+1}$, and another of the edge features, denoted as $\F^e_{k+1}$) using the method in \cite{zhang2018laser}, and stored in a buffer. Given the latest $M+1$ feature pointclouds from $\F_{w}$ to $\F_{k+1}$ ($w \triangleq k+1-M$), we will merge the first $M$ pointclouds from time $t_{w}$ to $t_{k}$ to construct the local map $\M_w \triangleq {}^{\fB_w}\M_w = (\M_{w}^p, \M_{w}^e)$ by using the latest estimated transforms $\hat{\tf}_{w}$, $\hat{\tf}_{w+1}$, $\dots$, $\hat{\tf}_{k}$. Then, for each feature pointcloud $\F_m$, $m = w, \dots k+1$, we will compute the set of feature coefficients $\Fcoef_m = \{\L^i_m \triangleq (\f^i, \n^i, c^i) \}$ that will be used to construct the Lidar feature factors. More details are discussed in Section \ref{sec: lidar feature factor}.


\begin{figure}
    \setlength\belowcaptionskip{-0.25cm}
    \centering
    \begin{overpic}[width=\linewidth,
                        ]{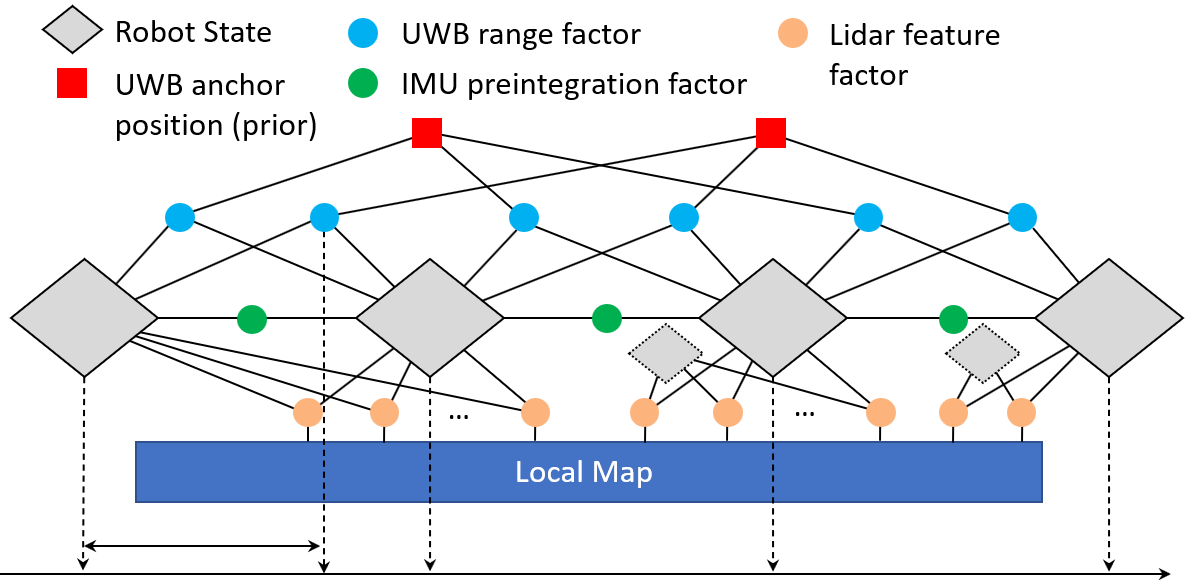}
                        \put(6.0,  -3.0){\footnotesize $t_w$}
                        \put(26.0, -3.0){\footnotesize $t^i$}
                        \put(15.0,  4.0){\footnotesize $\D t^i$}
                        \put(36.0, -3.0){\footnotesize $t_{w+1}$}
                        \put(65.0, -3.0){\footnotesize $t_{k}$}
                        \put(93.0, -3.0){\footnotesize $t_{k+1}$}
                        
                        \put(05.0,  21.82){\footnotesize $\X_w$}
                        \put(32.25, 21.82){\footnotesize $\X_{w+1}$}
                        \put(63.0,  21.82){\footnotesize $\X_{k}$}
                        \put(89.1,  21.82){\footnotesize $\X_{k+1}$}
                        
                        \put(54.0, 19.125){\scriptsize $\X_{w}$}
                        \put(80.5, 19.125){\scriptsize $\X_{w}$}
	\end{overpic}
	\vspace{0.01cm}
	\caption{Example of the factor graph over four time steps ($M=3$).
	Note that the time difference between the stamp of a UWB measuremnt and the latest preceding time step, denoted as $\Dt_i$ is also included in the UWB measurement $\U^i_m$.
	The factors are described in more details in Section \ref{sec: cost factor}.
    }
	\label{fig: factor graph}
\end{figure}
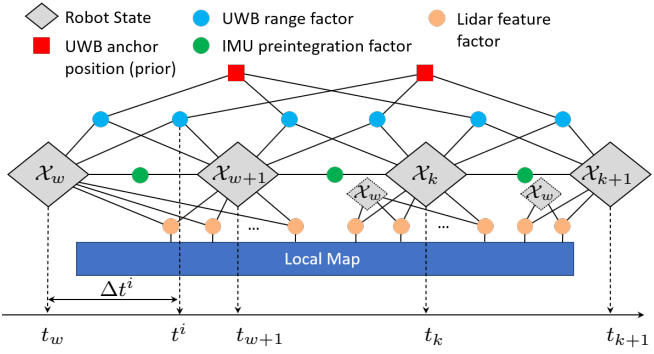

\subsection{Joint optimization sensor fusion}

Fig. \ref{fig: factor graph} illustrates the factor graph of our cost function, where the coupling between each UWB, IMU preintegration, and Lidar feature with the prior and the states is pictorially described. Thus, given all measurements from UWB, IMU and Lidar in the local window, the cost function can be constructed and optimized at time $t_{k+M}$ as
\begin{align}
    f(\hat{\X}) &\triangleq
    \Bigg\{\mkern9mu \sum_{m = k}^{k+M-1} \norm{\resi_\I(\hat{\X}_m, \hat{\X}_{m+1}, \I_m)}^2_{\vbf{P}_{\I_m}^{-1}} \nonumber\\
    &\qquad+ \sum_{m = k}^{k+M-1} \sum_{i = 1}^{N_\U^m}\norm{\resi_{\U}(\hat{\X}_m, \hat{\X}_{m+1} \U_m^i)}^2_{\vbf{P}_{\U_m^i}^{-1}} \nonumber\\
    &\qquad+ \sum_{m = k+1}^{k+M} \sum_{i = 1}^{N_\L^m}\norm{\resi_\L(\hat{\X}_{k}, \hat{\X}_{m}, \L_m^i)}^2_{\vbf{P}_{\L_m^i}^{-1}}\quad \Bigg\}, \label{eq: cost function}
\end{align}
where $r_{\I}(\cdot)$, $r_\U(\cdot)$, $r_\L(\cdot)$ are the \textit{residuals} constructed from UWB, IMU and Lidar measurements; $\vbf{P}_{\I_m}$, $\vbf{P}_{\U_m^i}$, $\vbf{P}_{\L_m^i}$ are the covariance matrices of the measurement error; $N_{\U}^m$ is the number of UWB measurements obtained in the period $(t_m, t_{m+1}]$, and $N_\L^m = \abs{\Fcoef_m}$. In this paper, we use the ceres solver\cite{ceres-solver} to optimize this cost function.


\section{Cost factors} \label{sec: cost factor}


\subsection{IMU preintegration factors:} \label{sec: IMU preint factor}

\subsubsection{IMU preintegration model}

For a conventional IMU, we can obtain the measurements $\breve{\angvel}$, $\breve{\accel}$, which are respectively the body's angular velocity and acceleration, corrupted by noises and biases. Given $\breve{\angvel}$, $\breve{\accel}$ and some nominal value of the IMU bias, denoted as $\breve{\bias}^{a}_{k}$, $\breve{\bias}^{\o}_{k}$, the IMU preintegration $\I_k \triangleq (\breve{\pia}_{k+1}, \breve{\pib}_{k+1}, \breve{\pig}_{k+1})$ can be calculated by:
\begin{align}
    \breve{\pia}_{k+1} &\triangleq \Int{t_k}{t_{k+1}}\Int{t_k}{u} {}^{k}\breve{\rot}_{s} (\breve{\accel}_s - \breve{\bias}^{a}_{k}) ds\ du, \label{eq: pIMU alpha noisy}
    \\
    \breve{\pib}_{k+1} &\triangleq \Int{t_k}{t_{k+1}} {}^{k}\breve{\rot}_{s} (\breve{\accel}_s - \breve{\bias}^{a}_{k}) ds,\label{eq: pIMU beta noisy}
    \\
    \breve{\pig}_{k+1} &\triangleq \Mint_{s=t_k}^{t_{k+1}} \breve{\pig}_{s} \circ \begin{bmatrix}0\\ \breve{\angvel}_s - \breve{\bias}_s^\o\end{bmatrix}, \label{eq: pIMU gamma noisy}
\end{align}
where ${}^k\breve{\rot}_{s} \triangleq \qtoR(\breve{\pig}_{s})$, and $\Mint_{s=t_k}^{t_{k+1}}(\cdot)$ denotes the integration of the quaternion derivative.
In practice, these integrations can be implemented by zero-order-hold (ZOH) or higher order methods (Runge-Kutta methods). The observation model $\I_k(\X) \triangleq(\pia_{k+1}(\X), \pib_{k+1}(\X), \pig_{k+1}(\X))$ can be stated as:
\begin{align*}
    &\breve{\pia}_{k+1} + \d\pia_{k+1} =
    \rot_k^{-1}(\pos_{k+1} - \pos_k - \vel_k\Dt_k + \frac{1}{2}{}\grav\Dt^2_k)
    \nonumber
    \\
    & - \vbf{A}_{{k+1}}^{\o} (\bias^{\o}_{k} - \breve{\bias}^{\o}_{k}) - \vbf{A}_{{k+1}}^{a} (\bias^{a}_{k} - \breve{\bias}^{a}_{k}) \triangleq \pia_{k+1}(\X_k, \X_{k+1}),
    \\
    &\breve{\pib}_{k+1} + \d\pib_{k+1} =
    \rot_k^{-1} (\vel_{k+1} - \vel_k + \grav\Dt_k)\dots
    \nonumber
    \\
    &- \vbf{B}_{{k+1}}^{\o}(\bias^{\o}_{k} - \breve{\bias}^{\o}_{k})
    - \vbf{B}_{{k+1}}^{a} (\bias^{a}_{k} - \breve{\bias}^{a}_{k}) \triangleq \pib_{k+1}(\X_k, \X_{k+1}),
    \\
    &\breve{\pig}_{k+1}
    \circ
    \begin{bmatrix} 1\\ \frac{1}{2}\vbf{C}_{k+1}^{\o}(\bias^{\o}_{k} - \breve{\bias}^{\o}_{k})\end{bmatrix}
    \circ
    \begin{bmatrix} 1\\ \frac{1}{2}\d\pith_{k+1}\end{bmatrix}
    \simeq
    \quat_k^{-1} \circ \quat_{k+1},
\end{align*}
where $\d\pia_{k+1}$, $\d\pib_{k+1}$, $\d\pith_{k+1}$ are the errors, whose covariance can be calculated via a propagation scheme, $\vbf{A}^{\o}_{k+1}, \vbf{A}^{a}_{k+1}, \vbf{B}^{\o}_{k+1}, \vbf{B}^{a}_{k+1}, \vbf{C}^{\o}_{k+1}$ are the Jacobians of the IMU preintegrations evaluated at the bias point $\bar{\bias}^{\star}_k$, i.e.
\begin{align*}
    &\vbf{A}_{{k+1}}^{\o}
    \triangleq
    \frac{\partial\breve{\pia}_{k+1}(\bar{\bias}_{k}^{\o})}
         {\partial \bar{\bias}^{\o}_{k}}
    \Big|_{\bar{\bias}_{k}^{\o} = \breve{\bias}_{k}^{\o}},
    \vbf{A}_{{k+1}}^{a}
    \triangleq
    \frac{\partial\breve{\pia}_{k+1}(\bar{\bias}_{k}^{a})}
         {\partial \bar{\bias}^{a}_{k}}
    \Big|_{\bar{\bias}_{k}^{a} = \breve{\bias}_{k}^{a}},
    \\
    &\vbf{B}_{{k+1}}^{\o}
    \triangleq
    \frac{\partial\breve{\pib}_{k+1}(\bar{\bias}_{k}^{\o})}
         {\partial \bar{\bias}^{\o}_{k}}
    \Big|_{\bar{\bias}_{k}^{\o} = \breve{\bias}_{k}^{\o}},
    \vbf{B}_{{k+1}}^{a}
    \triangleq
    \frac{\partial\breve{\pib}_{k+1}(\bar{\bias}_{k}^{a})}
         {\partial \bar{\bias}^{a}_{k}}
    \Big|_{\bar{\bias}_{k}^{a} = \breve{\bias}_{k}^{a}},
    \\
    &\vbf{C}_{k+1}^{\o}
    \triangleq 
    \frac{\partial\mathrm{Log}\l[{{}^k\breve{\rot}_{k+1}^{-1}(\breve{\bias}_{k}^{\o}){}^k\breve{\rot}_{k+1}(\breve{\bias}_{k}^{\o} + \bar{\bias}_{k}^{\o})}\r]}
         {\partial \bar{\bias}_{k}^{\o}}
    \Big|_{\bar{\bias}_{k}^{\o} = \zero}.
    %
\end{align*}

\subsubsection{IMU preintegration residual}

The IMU preintegration residual, denoted as $r_{\I}(\hat{\X}_k, \hat{\X}_{k+1}, \I_k)$ is therefore defined as
\begin{align}
    &r_{\I}(\hat{\X}_k, \hat{\X}_{k+1}, \I_k)
    \triangleq
    (
        r_{\pig},\
        r_{\pia},\ 
        r_{\pib},\ 
        r_{b}^{\o},\ 
        r_{b}^{a}
    ),
    \\
    &r_{\pig}
    \triangleq
    2\vbf{vec}
    \Bigg(
        \begin{bmatrix}
            1\\
            -\frac{1}{2}\vbf{C}_{k}^{\o}(\hat{\bias}^{\o}_{k} - \breve{\bias}^{\o}_{k})
        \end{bmatrix}
        \circ
        \breve{\pig}_{k+1}^{-1}
        \circ
        {}^{k}_{k+1}\hat{\quat}
    \Bigg),\\
    &r_{\pia}
    \triangleq
    \pia_{k+1}(\hat{\X}_k, \hat{\X}_{k+1}) - \breve{\pia}_{k+1},
    \\
    &r_{\pib}
    \triangleq
    \pib_{k+1}(\hat{\X}_{k}, \hat{\X}_{k+1}) - \breve{\pib}_{k+1},
    \\
    &r_{b}^{\o}
    \triangleq
    \hat{\bias}^{\o}_{k+1} - \hat{\bias}^{\o}_{k},
    \quad r_{b}^{a}
    \triangleq
    \hat{\bias}^{a}_{k+1} - \hat{\bias}^{a}_{k}, 
\end{align}
More comprehensive details on the IMU preintegration technique can be found at \cite{nguyen2020viral}.


\subsection{Lidar feature factors} \label{sec: lidar feature factor}

Recall that at time step $t_{k+1}$, we have a series of $M+1$ feature pointclouds $\F_w$, $\F_{w+1}$, $\dots$, $\F_{k+1}$, where the first $M$ are merged into a local map $\M_{w}$ consisting of points whose coordinates are \wrt to the body frame at time $t_w$ ($w = k+1-M$), i.e. the first pose in the local window. Given the feature pointcloud $\F^m$, we calculate the set of coefficients $\Fcoef_m$ following the steps in Algorithm \ref{algo: lidar coefficients}, which are indeed the parameters of the 3D planes that the corresponding feature points are supposed to belong to.

\begin{algorithm}
  \SetAlgoLined
  \KwIn{$\F_{m} = (\F_{m}^p,\ \F_{m}^e)$, $\M_w = (\M_{w}^p,\ \M_{w}^e)$, ${}^w_m\hat{\tf}$.}
  \KwOut{$\Fcoef_m = \{\L^i_m = (\f^i, \n^i, c^i) \}$}
  \For{$\mathrm{each}$ $\f \in \F_m$}{ \label{alg:alg1_corr_begin}
    Compute ${}^w\f$ from $\f$ using ${}^w_m\hat{\tf}$\;
    \uIf{ $\f \in \F_{m}^p$ }
    {
        Find $\mathcal{N}_\f = \text{KNN}({}^w\f,\ \M_w^p)$\;
        Find $\bar{\n} = \argmin_{\n} \sum_{\vbf{x} \in \mathcal{N}_\f} ||\n^\top \vbf{x} + 1||^2$\;
        Compute: $g = \frac{1}{\norm{\bar{\n}}}\l[1 - 0.9\frac{\abs{\bar{\n}^\top({}^w\f) + 1}}{\norm{\bar{\n}}\norm{{}^w\f}}\r]$\;
        Add $(\f, g\bar{\n}, g)$ to $\Fcoef_m$\;
    }
    \uElseIf{$\f \in \F_{m}^e$}
    {
        Find the set $\mathcal{N}_\f = \text{KNN}({}^w\f, \M_w^e)$, and its centroid $\bar{\pos} = \frac{1}{\abs{\mathcal{N}_\f}}\sum_{\vbf{x}\in\mathcal{N}_\f}\vbf{x}$\;
        Compute: $\vbf{A} \triangleq \frac{1}{\abs{\mathcal{N}_\f}}\sum_{\vbf{x}\in\mathcal{N}_\f}(\vbf{x} - \bar{\pos})(\vbf{x} - \bar{\pos})^\top$\;
        Find the eigenvector $\vel_{\max}$ corresponding to the largest eigenvalue of $\vbf{A}$\;
        Compute: $\vbf{x}_0 = {}^w\f$, $\vbf{x}_1 = \bar{\pos} + 0.1\vel_{\max}$, $\vbf{x}_2 = \bar{\pos} - 0.1\vel_{\max}$, $\vbf{x}_{01} = \vbf{x}_{0} - \vbf{x}_{1}$, $\vbf{x}_{02} = \vbf{x}_{0} - \vbf{x}_{2}$, $\vbf{x}_{12} = \vbf{x}_{1} - \vbf{x}_{2}$, \;
        Compute: $\n_1 = \vbf{x}_{12} \times (\vbf{x}_{10} \times \vbf{x}_{02})$,  $\n_1 \leftarrow \n_1/\norm{\n_1}$, $\n_2 = \vbf{x}_{12}\times\n_1$\;
        Compute: ${}^w\f_\bot = {}^w\f - (\n_1\n_1^\top)\vbf{x}_{01}$\;
        Compute: $c_1 = -\n_1^\top{}^w\f_\bot$ and $c_2 = -\n_2^\top{}^w\f_\bot$\;
        Compute: $g = \frac{1}{2}\l[1 - \frac{0.9\norm{\vbf{x}_{01}\times\vbf{x}_{02}}}{\norm{\vbf{x}_{12}}}\r]$\;
        Add $(\f, g\bar{\n}_1, g c_1)$ and $(\f, g\bar{\n}_2, g c_2)$ to $\Fcoef_m$\;
    }
  }
  \caption{Find Lidar feature coefficients}
  \label{algo: lidar coefficients}
\end{algorithm}

Hence, for each Lidar coefficient $\L_k^i = (\f^i, \n^i, c^i)$, a lidar feature factor can be constructed with the following residual:
\begin{align*}
    \resi_\L(\hat{\X}_w, \hat{\X}_k, \L_k^i) = (\n^i)^\top\hat{\rot}_w^{-1}\l[ \hat{\rot}_k\f^i + \hat{\pos}_k - \hat{\pos}_w\r] + c^i.
\end{align*}

\subsection{UWB range factors} \label{sec: uwb range factor}

Recall that in Section \ref{sec: framework}, for each interval $(t_{k},\ t_{k+1}]$ in the local window, we have a bundle of UWB measurements $\{\U^i_k\}_{i = 1}^{N^k_\U}$. More specifically, $\U^i_k$ is defined as:
\begin{align}
    \U^i_k &= \l(\breve{d}^i,\ \vbf{x}^i,\ \vbf{y}^i,\ \dt^i, \Dt_k\r),\ i = 1, 2, \dots N^k_\U,
\end{align}
where $\breve{d}^i$ is the distance measurement, $\Dt_k \triangleq t_{k+1} - t_{k}$ (see Fig. \ref{fig: factor graph}), $\dt^i \triangleq t^i - t_k$, $\vbf{x}^i$ is the position of the UWB anchor \wrt the world frame, and $\vbf{y}^i$ is the coordinate of the UWB ranging node in the body frame.

If we assume that the velocity and orientation of the robot change at a constant rate from time $t_{k}$ to $t_{k+1}$, then at time $t_{k} + \dt^i$ the relative position of a UWB ranging node $\vbf{y}^i$ from an anchor $\vbf{x}^i$ can be determined as:
\begin{align}
    {}^{\fW}\dis^i
    &= \dis(\X_{k}, \X_{k+1}\ \dt^i,\ \Dt_k) \nonumber \\
    &\triangleq \pos_{k+1} + \rot_{k}\Exp\l(s^i\Log(\rot_{k}^{-1}\rot_{k+1})\r) \vbf{y}^i \nonumber \\
    &\quad - \Dt_{k+1}\int_{s^i}^{1} \l[\vel_{k} + \tau(\vel_{k+1}-\vel_{k})\r]d\tau - \vbf{x}^i \nonumber \\
    &\triangleq \pos_{k+1} + \rot_{k}\Exp\l(s^i\Log(\rot_{k}^{-1}\rot_{k+1})\r) \vbf{y}^i \nonumber\\
    &\qquad\qquad\qquad\qquad\quad - a^i\vel_{k+1} - b^i{\vel}_{k} - \vbf{x}^i, \label{eq: ant-anc displacement}
\end{align}
where $s^i = \frac{\dt_i}{\Dt_k}$, $a_k^i = \frac{\Dt_k^2 - \dt_i^2}{2\D t_k}$, $b_k^i = \frac{(\Dt_k - \dt_i)^2}{2\Dt_k}$.

We consider the distance measurement $\breve{d}^i$ at time $t_k+\dt^i$ as the norm of the vector ${}^{\fW}\dis^i$, corrupted by a zero-mean Gaussian noise $\vbf{\eta}_{\U^i} \sim \mathcal{N}(0, \s_{\U}^2)$, i.e. $\breve{d}^i = \norm{{}^\fW\dis^i} + \vbf{\eta}_{\U^i}$. Thus, the UWB range factor can be defined as:
\begin{equation}\label{eq: uwb residual}
    \resi_{\U}(\hat{\X}_{k}, \hat{\X}_{k+1}, \U^i_k) \triangleq \| \dis(\hat{\X}_k, \hat{\X}_{k+1}, \dt^i, \Dt_k)\| - \breve{d}^i,
\end{equation}

\section{Experiments} \label{sec: experiment}

In this section we present the experiment results of the LIRO method on real world datasets. Video recording of the experiments can be viewed at \url{https://youtu.be/Wfp_VcwzNKY}, or the in the supplementary materials of this paper.

Fig. \ref{fig: hardware setup} presents the hardware setup for our experiments. Specifically, it consists of one VN-100 IMU, four UWB ranging nodes and two 16-channel OS1 Lidars, where one so-called horizontal Lidar is oriented to scan the surrounding effectively, and the other so-called vertical Lidar is set to scan the front, back and the ground effectively. These sensors are mounted on an hexacopter. The 3 anchors are deployed as depicted in Fig. \ref{fig: ranging scheme}, where the coordinate $x^*_1$ ranges from 40m to 60m and $y^*_2$ ranges from -15m to -10m, depending on the available space. The four UWB ranging nodes on the hexacopter are positioned on the vertices of a 0.75m $\times$ 0.55m rectangle around the IMU, which is also the body center. A Leica MS60 station with millimeter-level accuracy is used to provide groundtruth for the experiment. The software algorithms are implemented on the ROS framework\footnote{\url{https://www.ros.org/}}.

\begin{figure}[t]
    \setlength\belowcaptionskip{-0.25cm}
    \centering
    \includegraphics[width=0.9\linewidth]{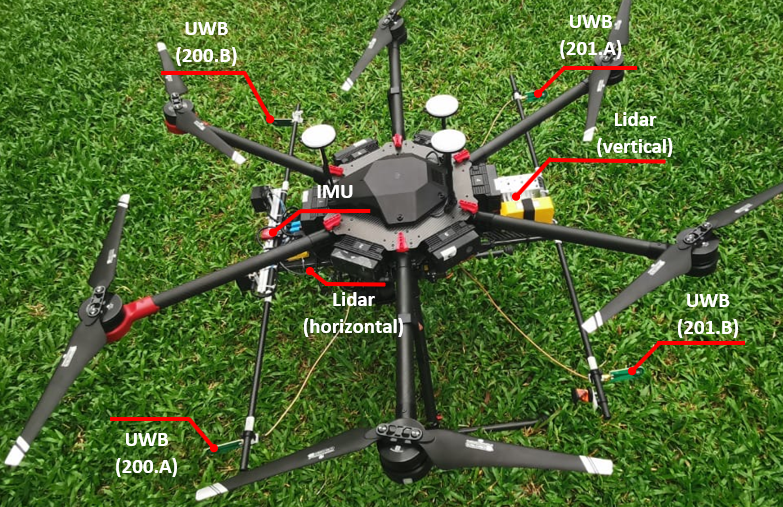}
	\caption{The hardware setup used in our flight tests.}
	\label{fig: hardware setup}
\end{figure}

We conduct three flight tests to scan the facade of a building to collect the data for these experiments. We run the LIRO algorithm on the horizontal and vertical Lidar measurements, and the ranging data being fused are limited to 0, 2, and 3 anchors, respectively, to demonstrate and compare the benefit of the number of landmarks in reducing estimation drift. We refer to the case when range data is not used as LIO and LIRO2 or LIRO3 to the cases where ranging to 2 or 3 anchors are fused, respectively. We use the method in \cite{zhang2018tutorial} to align the trajectories before calculating the positional root mean square error (RMSE) of the estimated trajectories. Though no groundtruth for orientation is available, we assume the hexacopter's onboard estimation, which fuses in the magnetometer measurement, is accurate and can be used as groundtruth, since no magnetic interference was observed at the test site on the day. This is then used to calculate the rotational RMSE.

\begin{figure}[t]
    \centering
    \includegraphics[width=\linewidth]{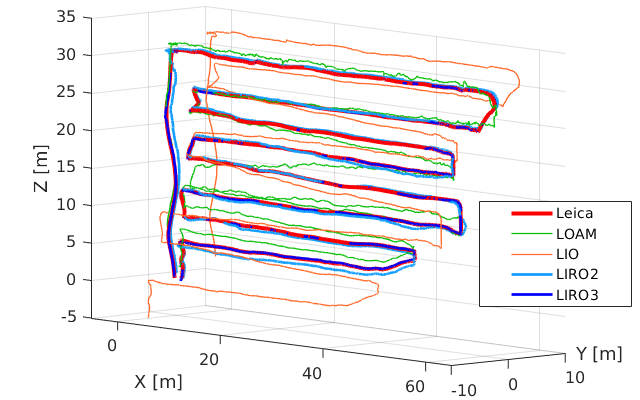}
	\caption{Trajectories of the estimates from the horizontal Lidar in the first test.}
	\label{fig: horz 321 traj}
\end{figure}

\begin{figure}[t]
    \centering
    \includegraphics[width=\linewidth]{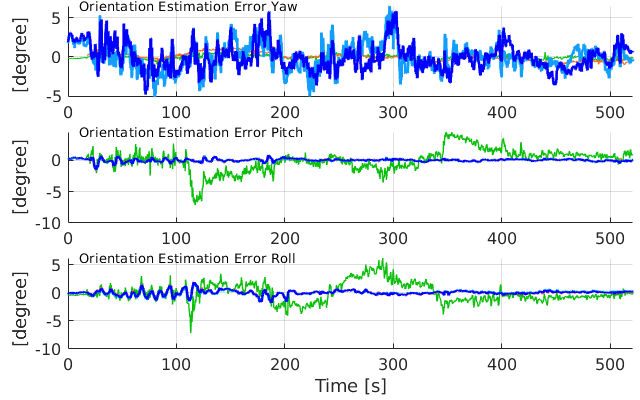}
	\caption{Orientation estimation error from the horizontal Lidar in the first test (the legend is the same with Fig. \ref{fig: horz 321 traj}).}
	\label{fig: vert 321 ypr err}
\end{figure}

The LOAM\footnote{\url{https://github.com/HKUST-Aerial-Robotics/A-LOAM}} method \cite{zhang2017low} is also run with the acquired data for comparison. We also attempt to run LIO-SAM \cite{shan2020liosam} with our datasets, however LIO-SAM quickly diverges after the robot takes off. We suspect this is because LIO-SAM requires the roll-pitch-yaw estimate from the IMU's built in estimator, which has significant drift due to high-frequency vibration when the hexacopter is flying. Indeed the experiment of LIRO with 0 anchors can also be considered as substitute for LIO-Mapping \cite{ye2019tightly}, since we adopt several software components from this work. However, it should be noted that our implementation has been significantly improved to ensure real-time performance, while LIO-Mapping is known to suffer from computation bottlenecks \cite{shan2020liosam}.
Tab. \ref{tab: horizontal ATE} and Tab. \ref{tab: vertical ATE} summarize the results of these experiments. Fig. \ref{fig: horz 321 traj} and Fig. \ref{fig: vert 321 ypr err} show the trajectory and orientation estimation error of some tests for a closer look. More plots and figures can be viewed in the accompanying video.
\begin{table}[t]
    \setlength{\tabcolsep}{2.5pt}
    \centering
    \renewcommand{\arraystretch}{1.1}
	\caption{Position and rotation RMSE of odometry estimates from the horizontal Lidar data.}
	\label{tab: horizontal ATE}
    \begin{tabu} to \textwidth {c||c|c|c|c||c|c|c|c}
    \hline\hline
                        &\mc{4}{c||}{$\text{RMSE}_\text{pos}$ [m]}
                        &\mc{4}{c}{$\text{RMSE}_\text{rot}$ [deg]}\\\cline{2-9}
    \mr{-2}{*}{Test} &LOAM &LIO &LIRO2 &LIRO3
                     &LOAM &LIO &LIRO2 &LIRO3\\\hline
    01 &1.499  &4.692       &0.527  &\bf{0.170}  
       &2.232  &\bf{0.693}  &1.650  &1.610\\     
    02 &5.275  &5.242       &0.758  &\bf{0.393}  
       &4.239  &\bf{0.927}  &2.351  &3.180\\     
    03 &2.549  &5.833       &0.664  &\bf{0.204}  
       &4.989  &\bf{0.996}  &2.597  &2.458\\     
    \hline\hline
    \end{tabu}
\end{table}

We can immediately see from Tab. \ref{tab: horizontal ATE} and Tab. \ref{tab: vertical ATE} that the use of ranging factors clearly improve the positioning estimation. While the RMSE of position estimate with only Lidar or Lidar-intertial measurements can be several meters large, by employing ranging to two anchors, the error can be reduced to below 0.8m, and the accuracy when using three anchors can be as small as 0.15m. Moreover, one notable feature that can be observed is that the use of ranging factors can regularize the quality of localization across the experiments. We can see that the RMSE is quite unpredictable in the LOAM and LIO cases, while the accuracy of LIRO is quite consistent across the environments and anchor configurations.

In terms of orientation estimate, we find that LIO estimation appears to be more accurate in most tests. We believe that this is due to LIO estimate being smoother than LIRO (refer to the yaw error in Fig. \ref{fig: vert 321 ypr err}). However it can be seen that these orientation errors are already quite small such that the difference is imperceptible. For example the largest difference between LIRO and LIO is approximately $2.3^o$ or 0.04 rad (test 02 in Tab. \ref{tab: horizontal ATE}), compared to a difference of several meters in the position errors. Moreover, since the accuracy of groundtruth is not guaranteed, the exact values of the RMSE are not be very important. On the other hand, the LIRO's orientation is of global type and the  accuracy is consistent throughout the experiments, while LOAM and LIO estimates are relative to the initial pose, and have unpredictable orientation drifts in some experiments.

\begin{table}[t]
    \setlength{\tabcolsep}{2.5pt}
    \centering
    \renewcommand{\arraystretch}{1.1}
	\caption{Position and rotation RMSE of odometry estimates from the vertical Lidar.}
	\label{tab: vertical ATE}
    \begin{tabu} to \textwidth {c||c|c|c|c||c|c|c|c}
    \hline\hline
                        &\mc{4}{c||}{$\text{RMSE}_\text{pos}$ [m]}
                        &\mc{4}{c}{$\text{RMSE}_\text{rot}$ [deg]}\\\cline{2-9}
    \mr{-2}{*}{Test} &LOAM &LIO &LIRO2 &LIRO3
                     &LOAM &LIO &LIRO2 &LIRO3\\\hline
    01 &13.665   &5.421      &0.492    &\bf{0.159}   
       &8.421    &1.812      &1.803    &\bf{1.639}\\ 
    02 &2.710    &2.033      &0.744    &\bf{0.394}   
       &4.864    &\bf{1.208} &2.463    &3.228\\      
    03 &1.603    &2.710      &0.643    &\bf{0.209}   
       &4.062    &\bf{0.801} &2.849    &2.786\\      
    \hline\hline
    \end{tabu}
\end{table}

\section{Conclusion} \label{sec: conclusion}

In this paper we have developed a tighly coupled Lidar-inertia-ranging odometry estimation scheme, so-called LIRO, and successfully implemented it on the open-source ceres solver and ROS. Experiments on real-world datasets have been conducted to verify the efficacy and effectiveness of the sensor fusion scheme. We show that via the use of some anchors that can be quickly deployed on the field, estimation drift can be significantly reduced, and the frame of reference for LIRO can be directly determined via the anchor deployment. The results demonstrate that LIRO is an effective and flexible localization solution for robots operating in semi-controlled environments.


\balance
\bibliographystyle{IEEEtran}
\bibliography{references}

\end{document}